# Multi-target Tracking of Zebrafish based on Particle Filter

CONG Heng[1,2], SUN Mingzhu[1,2], ZHOU Duoying[1,2], and ZHAO Xin[1,2]

1. Institute of Robotics and Automatic Information System, Nankai University, Tianjin 300071, China

2. Tianjin Key Laboratory of Intelligent Robotics, Nankai University, Tianjin 300071, China
E-mail: sunmz@nankai.edu.cn

**Abstract:** Zebrafish is an excellent model organism, which has been widely used in the fields of biological experiments, drug screening, and swarm intelligence. In recent years, there are a large number of techniques for tracking of zebrafish involved in the study of behaviors, which makes it attack much attention of scientists from many fields. Multi-target tracking of zebrafish is still facing many challenges. The high mobility and uncertainty make it difficult to predict its motion; the similar appearances and texture features make it difficult to establish an appearance model; it is even hard to link the trajectories because of the frequent occlusion. In this paper, we use particle filter to approximate the uncertainty of the motion. Firstly, by analyzing the motion characteristics of zebrafish, we establish an efficient hybrid motion model to predict its positions; then we establish an appearance model based on the predicted positions to predict the postures of every targets, meanwhile weigh the particles by comparing the difference of predicted pose and observation pose ; finally, we get the optimal position of single zebrafish through the weighted position, and use the joint particle filter to process trajectory linking of multiple zebrafish.

## 1 Introduction

Zebrafish, a vertebrate, is an excellent model organism. It is widely used in biology experiment [1], drug screening [2] and swarm intelligence [3] and so on due to its tenacious vitality, easy raising, and high homology with human genes. In recent years, there are a large number of techniques [4~6] for tracking of zebrafish involved in the study of behaviors, which makes it attack much attention of scientists from many fields.

It is well-developed for tracking of single fish but there are still a lot of challenges for tracking of multiple ones because of the complex motion. Firstly, it is hard to predict positions of zebrafish for the reasons of the high mobility and the uncertainty of motion. Secondly, it is difficult to establish a stable appearance model because of the similar appearances, fuzzy textures, and changeable shape of the fish bodies. Finally, the motion of zebrafish is independent in the three-dimensional space, but with the monocular camera top shooting, the motion will be projected to the two-dimensional space so that the trajectories appear crossing. And frequent trajectories crossing increases the difficulty of trajectory linking greatly.

In order to solve the problem above effectively, Reference [7, 8] detect and track multiple targets by linear prediction, but the methods often cause errors because of high mobility and randomness of zebrafish's motion. Reference [9] attempts to correct this error by using the zebrafish's texture features, but it consumes a large amount of time in sample calculation, and cannot be applied in the occasion where stable texture features cannot be captured.

In this paper, we find that the uncertainty of motion can be approximated by a set of random samples [10], which are captured in the state space by particle filter [11~14], and this method will be described in detail in the second part. To effectively predict the motion of zebrafish, we establish a hybrid motion model, which integrates an accelerated motion model and a visual inspection-based turning model, by analyzing the motion characteristics of zebrafish in 3.1. In 3.2, we establish an appearance model based on the predicted positions to predict the postures of every zebrafish, meanwhile weigh the particle by comparing the difference of predicted pose and observation pose, and then we use the weighted position to replace the optimal position of single zebrafish. In 3.3, we propose a joint particle filter method to deal with the problem of frequent occlusion when tracking multiple zebrafish. In the end, we do some experiments on tracking of multiple zebrafish in chapter 4.

## 2 Description of multi-target tracking of zebrafish

The task of this chapter is to describe the process of multi-target tracking of zebrafish, and introduce the particle filter, and then explain the meanings of involved symbols and formulas.

### 2.1 Description of the process of multi-target tracking of zebrafish

In order to track zebrafish effectively, the main work in this part is to estimate the maximum posterior probability of states at current time by using historical motion states and observation information. It is described as follows:

1) Using the states of historical positions and a motion model to predict all possible positions of zebrafish at current time t.
2) Predicting the zebrafish's poses on each predicted positions through the appearance model, and taking the matching degree between the predicted pose and the observed pose as the weight of each predicted state.

*This work is supported by National Natural Science Foundation of China under Grant 61273341 and 61327802. and the Major Science and Technology Project of Tianjin（14ZCDZGX00801）.

3) Listing all possible trajectory linking events when the trajectories of zebrafish separate from crossing images, and calculating the possibility of each trajectory linking event, and then taking the maximum one as the result of the trajectory linking.

## 2.2 Using Particle filter for multi-target tracking of zebrafish

First the state of zebrafish is defined as $X_{t-1}(l_x, l_y, a, b, \delta)$, including the predicted weight position $(l_x, l_y)$ and the optimal ellipse $(a, b, \delta)$, and observation information is defined as $Z_t(x, y, a', b', \delta')$, including the center (x,y) of zebrafish's image observed from the video sequence and the pose $(a', b', \delta')$ obtained by fitting the outer contour of zebrafish with ellipse. Here a(a') and b(b') represent the length and width of the zebrafish separately, and $\delta(\delta')$ represents the angle between the long axis of zebrafish's body and coordinate vector X axis.

The core of this work is to use particle filter for single zebrafish, as described below (1)(2), and to solve trajectories linking of multiple zebrafish after frequent occlusion, as shown below (3). The main method is described below:

1) Using the motion model $P(X_t | X_{1:t-1})$ to predict all possible positions $(l_x, l_y)^{(r)}$ of zebrafish, where r is the number of particles, according to the states of historical motion. It is studied that the state of zebrafish at current time is only related to the state at previous time, so it can be regarded as a Markov process, and the motion model can be simplified as $P(X_t | X_{t-1})$.

2) Predicting the posture $(a, b, \delta)$ of zebrafish through the appearance model on each predicted position $(l_x, l_y)^{(r)}$, and taking the matching degree between each predicted state $X_t^{(r)}$ and observed pose $Z_t$ as the weight of every predicted state, and finally using the weight position $\{X_t^{(r)}, \pi_t^{(r)}\}$ as the optimal position. The particle filter is as shown in formula (1):

$$P(X_t | Z^t) = cP(Z_t | X_t) \sum \pi_{t-1}^{(r)} P(X_t | X_{t-1}) \quad (1)$$

3) Using n particle filters $P(X_{it} | Z^t)(i = 1.2...n)$ for n zebrafish respectively. When there is $m(m \leq n)$ zebrafish separating from crossing images, it can be given m! trajectory linking events. After calculation of the sum of all weights of every trajectory linking event, the maximum one can be used as the result of the trajectory linking at the current time, as formula (2) shows.

$$P(X_t | Z^t) = MAX(P_k(\sum \pi_{it}^{(r)} | X_{it}, Z_t)), (k \in 1: m!) \quad (2)$$

## 3 Key technologies of multi-target tracking of zebrafish based on particle filter

This chapter mainly introduces the key technologies of multi-target tracking of zebrafish. Part 3.1 introduces the method for establishment of the motion model in detail. Part 3.2 establishes an appearance model and introduces the method of weighing the particles. In 3.3, a new method is proposed to deal with the problem for trajectory linking of zebrafish.

## 3.1 Prediction of the motion of zebrafish based on the motion model

In this part, a hybrid motion model is built, which combines the accelerated motion M and the visual inspection-based turning model R, through analyzing the statistics of motion velocity and turning characteristic of zebrafish.

First, the movement distance of zebrafish is predicted through the accelerated motion model M, and then the direction and turning angle are predicted by the visual inspection-based turning model R. Finally, the product of $R(\theta)M(L)$ is used to represent the variation of zebrafish's position at current time. The hybrid motion model is as follows:

$$(l_x, l_y)^{(r)} = P(X_t | X_{t-1}) = R(\theta)M(L) + X_{t-1} \quad (3)$$

### 3.1.1 Accelerated motion model M

The main task of the accelerated motion model M is to predict the movement distance $L_t$ of zebrafish through historical positions at current time t. In order to effectively predict the mutation of the velocity and the acceleration in the process of zebrafish's motion, a three-order accelerated motion model is introduced as the basic framework, in which w(t) is the Gauss white noise, and the three-order accelerated motion model is as follows:

$$\begin{bmatrix} \dot{L}_t \\ \ddot{L}_t \\ \dddot{L}_t \end{bmatrix} = \begin{bmatrix} 0 & 1 & 0 \\ 0 & 0 & 1 \\ 0 & 0 & 0 \end{bmatrix} \begin{bmatrix} L_{t-1} \\ \dot{L}_{t-1} \\ \ddot{L}_{t-1} \end{bmatrix} + \begin{bmatrix} 0 \\ 0 \\ 1 \end{bmatrix} \omega(t) \quad (4)$$

The movement distance $L_t$ of zebrafish can be calculated through the predicted velocity $(\dot{L}_t)$, acceleration $(\ddot{L}_t)$, and jerk $(\dddot{L}_t)$. But the randomness of zebrafish's positions cannot be described in the motion process, that is, it is impossible for the zebrafish to completely follow the ideal motion model. So statistical trajectory information of zebrafish is used to generate multiple predicting distances.

The analysis of the trajectory information of zebrafish reveals a motion law that the movement distances generally conform to the Gaussian function, where $L_x$ denotes the mean value of movement distances is, and $\sigma_v$ denotes the variance. The statistical method is shown in experiment 4.1. The predicted movement distance $L_t$ can be used to replace the mean value $L_x$ in L to generate multiple predicting distances by the three-order accelerated motion model. The formula is shown as follows:

$$M(L) = N(L_t, \sigma_v^2) \quad (5)$$

### 3.1.2 Visual inspection-based turning model R

The main task of the visual inspection-based turning model R is to predict the direction and the turning angle according to the motion characteristics of zebrafish. Here, the rotation matrix is used as the basic framework to design the turning model. The experiments show that there are two characteristics for the motion of zebrafish.

1) As zebrafish turns, the value of turning angle is gradually attenuated, and it is only related to the turning angle at the previous time.

2) Only when the body of zebrafish is bent, it can obtain power for turning motion, and the direction is on the side of the bent fish body's gravity center.

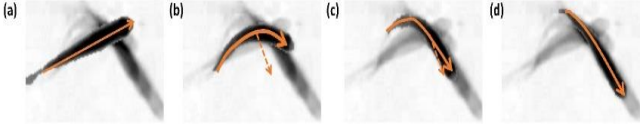

Fig.1: Schematic diagram of zebrafish's turning motion

Fig.1 shows four consecutive frames for the turning motion of zebrafish. Overlaying these four images and blurring the foreground object of the resulting image, the motion trace can be reflected. The turning angle is continuously attenuated when zebrafish turns. In Fig.1 (b), the body of zebrafish is bent to provide power for the turning motion, and the direction of the turning is on the side of the fish body's center of gravity.

The direction and the turning angle of zebrafish can be obtained through the two characteristics above. Firstly, the turning angle $\theta_{t-1} = |\delta_{t-1} - \delta_{t-2}|$ is defined by using the absolute subtraction value of the angle for one fish body in two consecutive frames, and then the product of the turning angle $\theta_{t-1}$ and the attenuation constant d is taken as the prediction of the zebrafish's turning angle at time t. Here, the direction of zebrafish's center of gravity is obtained by image inspection, and $\pm$ is used to indicate the turning direction. The formula of turning angle can be expressed as follows:

$$\theta_t = \pm d(\delta_{t-1} - \delta_{t-2}) \quad (6)$$

Also in order to describe the randomness of turning angle of zebrafish, the same method mentioned above is applied to obtain the probability distribution of turning angles of zebrafish. The statistical results show that the probability distribution is $\theta = N(0, \sigma_\theta^2)$. The predicted turning angle $\theta_t$ is used instead of mean value 0, formula as follows:

$$R(\theta) = N(\theta_t, \sigma_\theta^2) \quad (7)$$

The positions of zebrafish can be predicted by the two models. Fig.2 is a schematic diagram of the hybrid motion model, where the blue oval area is the possible positions predicted by the hybrid motion model, and its internal state distribution is the joint distribution of $R(\theta)$ and $M(L)$. And its state probability density reflects the randomness of the motion of zebrafish.

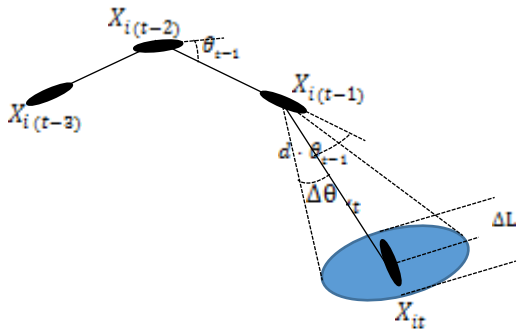

Fig. 2: Schematic diagram of the hybrid motion model

### 3.2 Weighing particles based on appearance model

The task of this part is to predict the poses of zebrafish based on the positions predicted above, and to take the matching degree between each predicted pose and observed pose as the weight of every predicted state. This part first describes the observation information, and then introduces the appearance model, and in the end, describes the method of particles' weights computing.

#### 3.2.1 Observation information and appearance model of zebrafish

Observation $Z_t(x, y, a', b', \delta')$ contains the center position (x,y) of the observed image and the ellipse $(a', b', \delta')$ which is used to fit the outer contour of zebrafish through the principal component analysis method (PCA). This ellipse can reflect the direction of zebrafish even when the body is bent, as shown in Fig.3.

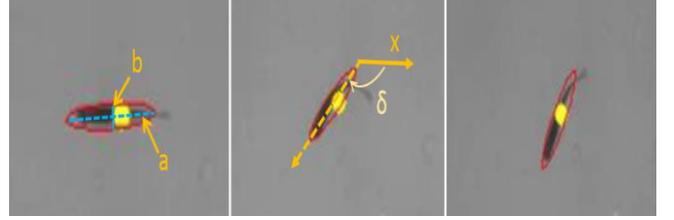

Fig.3: Fitting outer contour of zebrafish with ellipse through PCA

Ellipse is used as the appearance model of zebrafish. First the predicted positions $(l_x, l_y)^{(r)}$ are used as the center of ellipse. Then, for determining the elliptical inclination angle, which is the core of the appearance model, the sum of optimal inclination angle $\delta_{t-1}$ in the previous frame and the turning angle $\theta_t$ in formula (6) is taken as the mean value of inclination angles at current time. Combined with the variance $\sigma_\theta$ of turning angles mentioned above, a plurality of inclination angles $\delta$ are generated through the Gaussian function, as shown in formula (7):

$$\delta = N(\delta_{t-1} + \theta_t, \sigma_\theta^2) \quad (7)$$

When there is no targets interaction in zebrafish school, the nearest neighbor method can be used to find the ellipse in each observation information as the appearance of the current state, and update the average value of lengths $\bar{a}$ and width $\bar{b}$. Nevertheless, when there is some targets interaction in zebrafish school, the predicted pose $(\bar{a}, \bar{b}, \delta)$ should be used as the appearance of the state, with the particles expressed as $U_t^{(r)}(l_x, l_y, \bar{a}, \bar{b}, \delta)$.

#### 3.2.2 Calculation of Particles' Weights

Predicted poses $U_t^{(r)}(l_x, l_y, \bar{a}, \bar{b}, \delta)$ can be generated through the above-mentioned method. First the weight of each predicted state is computed, and then the weights should be normalized. In the end, the weighted position is used as the optimal position at current state, and the ellipse with the maximum weight is regarded as the optimal appearance.

The method of computing weights is described as follows. The foreground pixel of the observation $Z_t$ is labeled as F=1, while the background pixel is labeled as F=0. W represents the number of the foreground pixels F=1 contained in the current predicted ellipse $U_t^{(r)}(l_x, l_y, \bar{a}, \bar{b}, \delta)$, while S denotes the area of the predicted ellipse. After defining these parameters

and symbols, each particle's weight can be computed. The formula is as follows:

$$\pi_t^{(r)} = \frac{W(Z_t | U_t)}{S} \quad (8)$$

### 3.3 Trajectory linking based on joint particle filter

The established hybrid motion model and the appearance model can form a particle filter, and n particle filters are used for n zebrafish to track the trajectories of n zebrafish. However, given that frequent occlusion of zebrafish, many particle filters may track the same target when zebrafish separates from crossing images. Therefore, a joint particle filter is proposed to assume all possible trajectory linking events, selecting the most likely assumption as the trajectory linking at current time.

#### 3.3.1 Analysis of types of targets interaction

First all possible assumptions are enumerated. Given that the number of target is fixed, each separated zebrafish must correspond with each relevant observation, that is, m! assumptions of trajectory linking events will be obtained when m targets separate from the crossing images.

The second step is to calculate the probability of each assumption, using the maximum one as the trajectory linking at current time. As shown in Fig.4, two particle filters $X_{A(t-1)}$ and $X_{B(t-1)}$ get crossing before time t-1, and then it can be found that the targets separate from the crossing images at time t, and meanwhile two observation information $Z_{Ct}$ and $Z_{Dt}$, which will generate two assumptions of trajectory linking events, P1 and P2 (P1, A-C, B-D; P2, A-D, B-C), are detected. In addition, both particle filters $X_{A(t-1)}$ and $X_{B(t-1)}$ will predict particles on both observation images, so the sums of all particles' weights in each assumption are separately taken as the probability of the current assumption. The specific method is as follows:

1) For assumption P1, all weights' sum $\sum \pi_t^{AC}$ of particles generated by the particle filter $X_{At}$ in the observation $Z_{Ct}$ and all weights' sum $\sum \pi_t^{BD}$ of particles generated by the particle filter $X_{Bt}$ in the observation $Z_{Dt}$ should be calculated.

2) For assumption P2, all weights' sum $\sum \pi_t^{AD}$ of particles generated by the particle filter $X_{At}$ in the observation $Z_{Dt}$ and all weights' sum $\sum \pi_t^{BC}$ of particles generated by the particle filter $X_{Bt}$ in the observation $Z_{Ct}$ should be calculated.

3) By comparing assumption P1's value ($\sum \pi_t^{AC} + \sum \pi_t^{BD}$) and assumption P2's value ($\sum \pi_t^{AD} + \sum \pi_t^{BC}$), the assumption with the maximum value is selected as the trajectory linking at the current time.

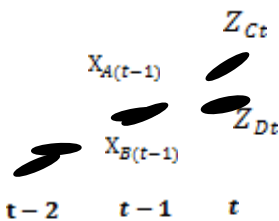

Fig.4: Schematic Diagram of Multiple Hypotheses

## 4 Results of experiments on tracking of zebrafish

The experiments test five swimming zebrafish in the video provided by Reference [9]. The light source is arranged on the top of the water tank so that the camera can clearly shoot contours of zebrafish. The video's frame rate is 15 frames /s. This chapter first presents the statistical method which obtains the parameters of zebrafish, and then analyzes the performance of hybrid motion model, and finally, applies the joint particle filter to track five zebrafish and analyze the experimental results.

### 4.1 Data statistics of trajectories zebrafish

In part 3.1 and 3.2, we established a hybrid motion model and an appearance model by using Gaussian function. The statistical method of the movement distances and turning angles of zebrafish is described as follows:

1) Obtaining the trajectory of zebrafish through the target tracking software, and compute the velocity and the turning angle between each adjacent frames.

2) Establishing velocity statistical histogram, in which the movement distance is taken as the abscissa and the unit pixel is taken as the coordinate interval.

3) Establishing turning angle statistical histogram, in which the turning angle is taken as the abscissa and the unit angle is taken as the coordinate interval.

4) Using the Gaussian function to fit the velocity statistical histogram and the turning angle statistical histogram.

According to the experimental result, the motion velocity follows the statistical distribution $L = N(L_x, 3.885^2)$, and the turning angle statistical distribution accords with the double Gaussian distribution, with the variance $\sigma_{\theta 1} = 1.478, \sigma_{\theta 2} = 7.271$. It also can be found that 90% of the turning angle's values are within $\pm 15^o$.

### 4.2 Effect of the hybrid motion model

This part first proves that the hybrid motion model can adapt to the high acceleration and fast turning characteristics of zebrafish, and then analyzes the overall error parameters of the hybrid motion model.

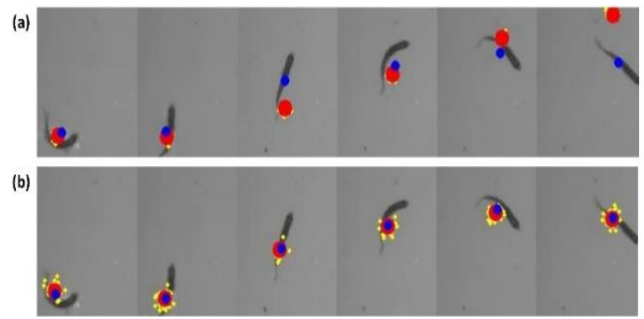

Fig.5: Comparison of effects of the hybrid motion model and the uniform motion model

Fig.5 represents separately using the uniform motion model (see Fig.5a) and the hybrid motion model (see Fig.5b) to predict the motion of zebrafish. The first 3 frames are about zebrafish doing accelerated motion, and the last 3 frames are about zebrafish doing turning motion. The blue spot are the observation positions, the red spot is the area predicted by the motion model, and the yellow spots are the particles. It is easy to find that the uniform motion model cannot keep tracking when acceleration and direction of zebrafish is changed, which is also the important reason that the uniform motion model cannot effectively track zebrafish smoothly in other methods, while the hybrid motion model can effectively keep tracking of zebrafish in the whole process.

To analyze the performance of hybrid motion model, we select 1000 consecutive frames to get the data of tracking, and the statistical graphs of velocity, acceleration, turning angle and deviation error in zebrafish's motion were separately drawn, with frame as the abscissa and pixel distance as the ordinate, as shown in Fig.6. By analyzing the data, it is found that there is a large deviation in the motion model when there is high acceleration and large-angle turning in the zebrafish's motion, which further reflects the complexity of the zebrafish's motion. Also it can be seen that the hybrid motion model can be used to track zebrafish in most cases.

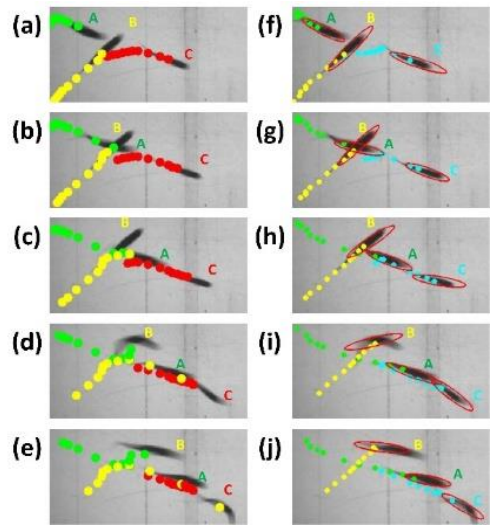

Fig.7: The appearance model's correction to the trajectories

### 4.4 Multi-target tracking of zebrafish based on the joint particle filter

In the part 3.4, Joint Particle Filter is presented to track the trajectories of multiple zebrafish. Now, for tracking five zebrafish, it is stipulated that when the deviation between the position predicted by the particle filter and the observed position is too large, the trajectory linking error happens, and the program will automatically record position error and take the wrong position as the initial position. The experimental parameters are set as follows: the variance of motion velocities in the accelerated motion model is $\sigma_l = 3.885$, the variances of turning angles in the turning model are $\sigma_{\theta 1} = 1.478, \sigma_{\theta 2} = 7.271$. Some tracking results are shown in Fig.8.

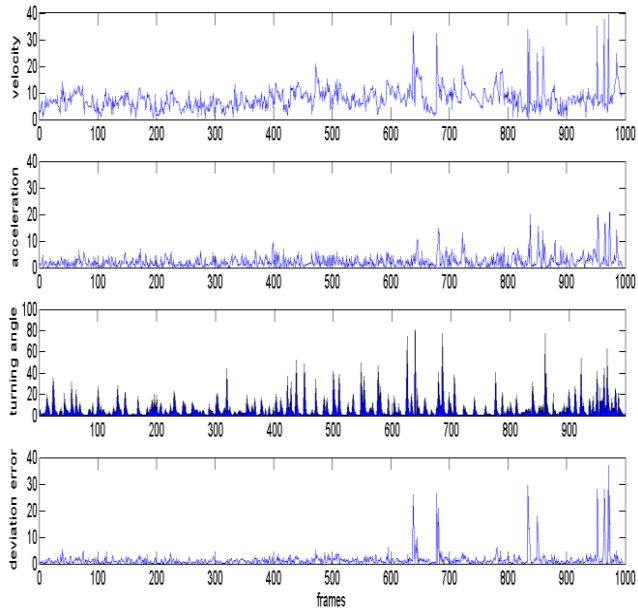

Fig.6: Error analysis of the motion model

### 4.3 The appearance model's correction to the trajectories

In the part 3.2, ellipse was used as the appearance model of zebrafish. The appearance model cannot only effectively calculate the weight of particles, but also can correct the wrong trajectory linking, and to a certain extent, distinguish the position of zebrafish in the crossing image. As shown in Fig.7, in the left column (a~e), trajectory tracking only uses zebrafish's position information, causing that the errors in the trajectory linking cannot be avoided; in the right column (f~j), the ellipse is used as the appearance model, leading to the right trajectory linking even when there is some occlusion.

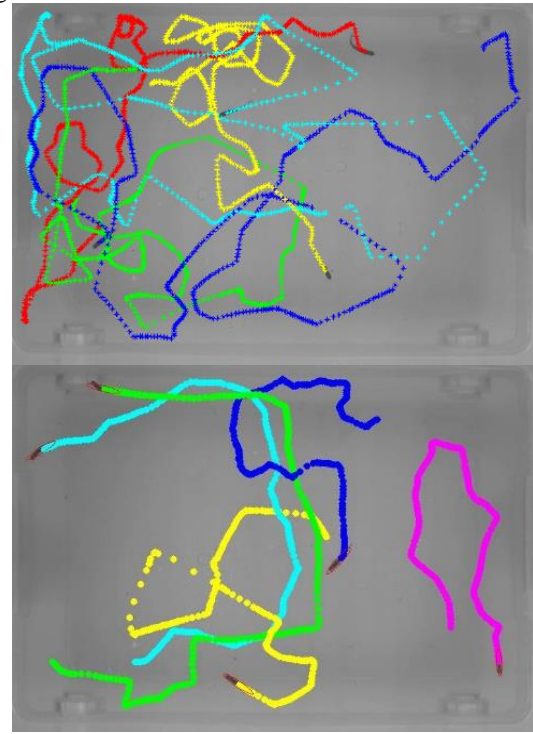

Fig.8: Tracking five zebrafish at the same time by the joint particle filter

Five zebrafish can be tracked, and semi-automatically the wrong trajectory linking can be corrected according to the records, causing correctly tracking zebrafish in the whole process.

## 5 Conclusion

Through the establishment of hybrid motion model, appearance model and joint particle filter. The main contributions of this paper are:

1) Presenting a method which obtains the motion parameters of zebrafish efficiently, and the statistical theory can be applied to other tracking methods.

2) Proposing a hybrid motion model that can adapt to the mobility and the uncertainty of zebrafish.

3) Distinguishing the position of zebrafish from the crossing image by an appearance model.

4) Using the multiple hypotheses method in the joint particle filter to effectively deal with the trajectory linking produced by occlusion.

In the future research, we will aim at realizing automatically obtaining the motion parameters of zebrafish during the tracking process, changing the parameters' values according to the different individual, and taking the parameters as a characteristic to correct the wrong trajectory linking. On the other hand, we will consider using the binocular camera to replace the monocular camera, which can avoid losing a lot of motion information, and applying the joint particle filter to the three-dimensional motion trajectories obtained by the binocular camera, reducing the error rate of trajectory linking.